\documentclass[conference]{IEEEtran}
\usepackage{array}

\usepackage[utf8]{inputenc}
\usepackage{amsmath, amssymb}
\usepackage{graphicx}
\usepackage{cite}
\usepackage{hyperref}

\usepackage{breqn}
\usepackage{lipsum}
\usepackage{multirow}
\usepackage{algorithm}
\usepackage{algpseudocode}
\usepackage{lmodern}

\title{Multi-Relation Extraction in Entity Pairs using Global Context}

\author{
    \IEEEauthorblockN{Nilesh, Atul Gupta, Avinash C Panday}
    \IEEEauthorblockA{
        Department of Computer Science and Engineering \\
        PDPM IIITDM Jabalpur, India \\
        \{nilesh, atul, avish.p\}@iiitdmj.ac.in
    }

}

\begin{document}

\maketitle

\begin{abstract}
In document-level relation extraction, entities may appear multiple times in a document, and their relationships can shift from one context to another. Accurate prediction of the relationship between two entities across an entire document requires building a global context spanning all relevant sentences. Previous approaches have focused only on the sentences where entities are mentioned, which fails to capture the complete document context necessary for accurate relation extraction. Therefore, this paper introduces a novel input embedding approach to capture the positions of mentioned entities throughout the document rather than focusing solely on the span where they appear. The proposed input encoding approach leverages global relationships and multi-sentence reasoning by representing entities as standalone segments, independent of their positions within the document. The performance of the proposed method has been tested on three benchmark relation extraction datasets, namely DocRED, Re-DocRED, and REBEL. The experimental results demonstrated that the proposed method accurately predicts relationships between entities in a document-level setting. The proposed research also has theoretical and practical implications. Theoretically, it advances global context modeling and multi-sentence reasoning in document-level relation extraction. Practically, it enhances relationship detection, enabling improved performance in real-world NLP applications requiring comprehensive entity-level insights and interpretability.
\end{abstract}

\section{Introduction}~\label{Introduction}
\begin{figure*}[htbp!]
	\centering 
	\includegraphics[width=0.65\textwidth]{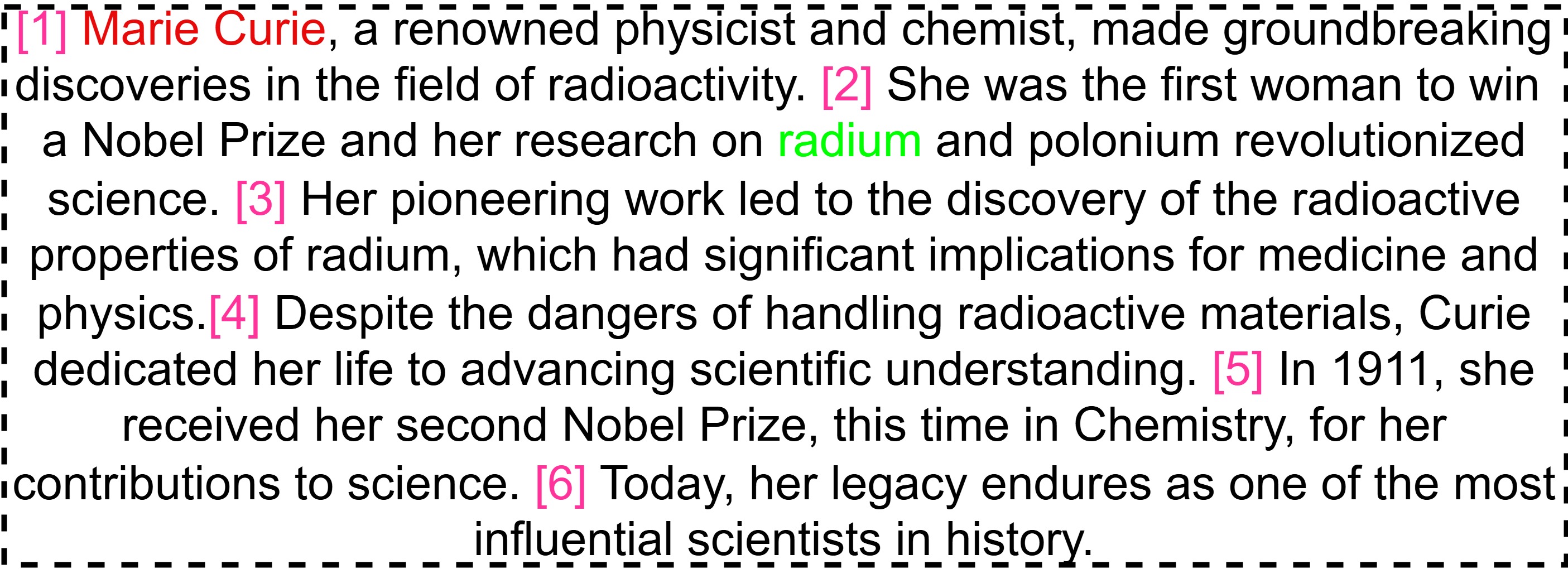}	
	\caption{Example of DRE} 
	\label{Example of DRE}
\end{figure*}

Relation extraction involves identifying the semantic relationship between two entities in the text span \cite{pawar2017relation}. Relation extraction can be performed at two levels: sentence level and document level. In sentence-level relation extraction, relationships between two entities within a single sentence are identified. On the contrary, document-level relationships require a different approach to handle cross-sentence implicit connections \cite{delaunay2023comprehensive,zhao2024comprehensive},. These implicit connections involve indirect relationships that require inferring the connection between entity pairs by reasoning over the surrounding information of entities in the document \cite{sun2022dual}. For example, ``Marie Curie conducted groundbreaking research in radioactivity. Her work changed the understanding of radium." The relationship between "Marie Curie" and "Radium" is implicit and requires inferring the contextual reasoning that describes her work on radioactivity. Document-level relation (DRE) has unique implicit connection challenges. First, extracting the relation between entity pairs in cross-sentence is more complex where the relationship is not mentioned in the document. For example, as shown in Figure \ref{Example of DRE}, the relation between ``Marie Curie" and ``Radium" is not directly mentioned in the document. Therefore, it is essential to address cross-sentence dependencies and establish context-aware relationships effectively. Second, treating all entity references equally can weaken representation, as entities may appear multiple times throughout the document in different forms. For instance, ``Marie Curie," ``Curie," and ``Her legacy" appear in lines 1, 5, and 6 of Figure \ref{Example of DRE}, respectively, highlighting the need for proper entity disambiguation. Third, making connections or drawing inferences among entities such as ``Marie Curie," ``radium," "polonium," ``Nobel Prize," ``science," ``medicine," and ``physics" often requires multiple steps or intermediary links, particularly for relationships like ``achievements," which must be inferred by examining the document as a whole. The long-distance \cite{jia2024document}, multi-hop reasoning \cite{zhao2024corex}, where certain relationships are not explicitly mentioned, relies on indirectly connecting entities throughout the text.

Numerous studies in the literature have attempted to address the challenges associated with document-level relation extraction. Most existing solutions construct document graphs using knowledge-enhanced representations \cite{sahu2019intersentencerelationextractiondocumentlevel} combined with heuristic rules, allowing them to capture cross-sentence dependencies and perform long-distance reasoning. The graph-based methods demonstrated improved performance for sentence-level relation extraction; however, they often failed to identify relationships between entities at the document level. Additionally, Transformers, viewed as graph neural networks with multi-head attention, are utilized to handle long-distance dependencies. Some other works, such as \cite{Zhou_Huang_Ma_Huang_2021} and \cite{Xu_Wang_Lyu_Zhu_Mao_2021}, have employed pre-trained models for document relation extraction (DRE) without relying on graph structures, instead applying average or max pooling to identify coreferential entities.

The above discussion highlights the significance of relation extraction in natural language processing, particularly in identifying semantic relationships between entities. Document-level relation extraction (DRE) presents unique challenges, such as cross-sentence dependencies and the necessity for long-distance reasoning. In this context, implicit connections require analyzing the entire document rather than isolated sentences. Despite recent advancements, DRE remains complex due to referential entities and indirect connections. Therefore, this paper introduces a novel method to enhance the efficacy of the DRE model. The proposed method processes the entire document by treating each entity pair individually rather than simply grouping all mentioned entities, as done in prior work \cite{Zhou_Huang_Ma_Huang_2021}. This approach ensures more explicit recognition of global context and cross-sentence relationships throughout the document. The model captures the appropriate relationships more effectively by isolating each entity pair alongside the document's global context. Moreover, graph-based reasoning networks are employed to facilitate more advanced reasoning. The model uncovers useful patterns and insights that emerge from broader contexts by leveraging global interactions across all entity pairs. Furthermore, the intrinsic dependencies and correlations between various entity pairings further enhance the model's reasoning capabilities. In the proposed method, an input-encoded component has been introduced to capture the global context of a document by feeding separate entity pairs into the BERT encoder.  The major contributions of the paper are summarized below-
\begin{itemize}
    \item A new input embedding method leveraging global relationships and multi-sentence reasoning has been introduced. In the proposed approach, entities are treated as self-contained segments that are distinct from the main text, ensuring each entity remains independent of the positions where they are mentioned.
        
    \item The classification layer, optimized initially for 96 relations, is customized to efficiently map the classification token representation to fine-grained relation categories that can be generalized for any number of relations based on the training dataset. 
    
    \item The proposed method has been validated on the three benchmark document-level relation extraction datasets. The experimental outcomes affirm the efficacy of the proposed approach.
\end{itemize}

\section{Related Work}

Relation extraction is essential in various natural language processing (NLP) tasks, especially knowledge graph construction. Initially, research in this domain was confined to sentence-level relation extraction (SRE), which focuses on predicting relations between two entities within a single sentence. Several methods have been introduced in the literature for sentence-level relation extraction using traditional relation extraction methods \cite{zeng2014relation}, \cite{cai2016bidirectional} and more recently, Zheng et al. \cite{zheng2021prgc} illustrated how pre-trained and fine-tuned models can address this task. However, entity pairs may appear in multiple sentences throughout the documents, which may restrict the SRE in practice, where the relationship depends on the overall context of various sentences. Therefore, research shifted from SRE to document-level relation extraction (DRE) \cite{yao2019docred}. In DRE, the same entity may appear multiple times across different sentences, creating a need to capture multi-sentence context for complex DRE reasoning. The DRE task typically involves classifying all possible pairs of entities in a document and assigning multiple relations to each pair to capture complex relationships (\cite{zhou2021document}. Recently, \cite{christopoulou2019connecting} introduced a novel DRE method that first learns contextual representations of the document and its tokens, subsequently constructs entity representations using a BERT-based model, and finally employs a classifier for multi-label classification.

Document representation can be approached in two primary ways. The first method, presented by Quirk and Poon \cite{quirk2016distant}, involves designing a document graph where new dependencies are introduced through adjacent sentences and discourse relations. The second, proposed by \cite{christopoulou2019connecting}, constructs a document graph using diverse node and edge types. Meanwhile, Nan and Lu \cite{nan2020reasoning} pioneered a document-level approach to relation extraction that generates latent dependency trees on the fly, treating them as dynamically formed structures within the document. Several studies, including \cite{wang2020global}, \cite{zeng2020double}, \cite{li2020graph}, and \cite{zhang2020document}, have focused on improving models by incorporating structural dependencies and employing graph-based algorithms for implicit reasoning. However, these models encountered challenges in effectively handling implicit reasoning and capturing complex relationships and dependencies. Therefore, Zeng et al. \cite{zeng2020double}, Li et al. \cite{li2021mrn}, and Xu et al. \cite{xu2021document} introduced specialized reasoning architectures to handle relation inference more effectively, thereby improving both accuracy and clarity in reasoning. Furthermore, zhou et al. \cite{zhou2021document} advanced the field by introducing adaptive thresholding loss and localized context pooling to enhance performance. Likewise, \cite{xu2021entity} incorporated entity structure dependencies into the transformer encoding process to capture complex relationships. Ru et al. \cite{ru2021learning} proposed LogiRE, a probabilistic model consisting of a rule generator and relation extractor to address the challenge of learning logical rules for relation extraction. The proposed method showed better performance for relation extraction. However, they perform poorly if the document comprises noisy data. Thus, Zhang et al. \cite{zhang2021modular} introduced a model to address the noise and reliability issues in the learning process by dividing the document-level relation extraction task into two sub-problems: relation detection and argument resolution, leading to more reliable outcomes.

Based on the above discussion, it has been observed that existing methods primarily focus on local or pairwise relations between entities to extract relationships at the document level \cite{zhou2021document}, \cite{zhang2021modular}. Due to this, their performance generally degrades if multiple relations exist between the same entities across the document. Therefore, in this paper, a novel method has been introduced. Unlike earlier approaches, the proposed method employs an input encoding strategy to emphasize global document context. The proposed modification overcomes the constraints of earlier models by capturing rich contextual information without relying exclusively on local or pairwise relations. By integrating global context directly into the encoding process, the proposed method enhances the model’s ability to reason about relationships across the entire document.

\section{Methodology}
\label{Methodology}
This section presents the proposed method, which integrates input embedding based on global relationships and multi-sentence reasoning to achieve more contextually accurate DRE predictions. The proposed method consists of three primary components: (i) input encoding, (ii) BERT encoder layers, and (iii) a classification layer. The following subsection presents the details of the proposed method. In addition, the detailed steps of the algorithm are depicted in Algorithm \ref{algo:1}.

\begin{algorithm}[ht!]
\caption{DocLevel-RE}
\scriptsize
\label{algo:1}
\begin{algorithmic}[1]
\Require $X = \text{concatenate}(\text{[CLS]}, T, \text{[SEP]}, H_e, \text{[SEP]}, T_e, \text{[SEP]})$
\Ensure Predicted relation $r_{pred}$

\State $X_{\text{emb}} \gets E_w(X) + E_p(\text{Pos}(X)) + E_t(\text{Type}(X))$
\State $H^0 \gets X_{\text{emb}}$
\For{$i = 1$ to $12$}
    \State $H' \gets \text{MHA}(H^{i-1}) + H^{i-1}$
    \State $H'' \gets \text{LayerNorm}(H')$
    \State $H^i \gets \text{FFN}(H'') + H''$
    \State $H^i \gets \text{LayerNorm}(H^i)$
\EndFor
\State $C \gets \tanh(W_p \cdot H^N_0 + b_p)$
\State $R \gets \text{Dropout}(C)$
\State $P(R) \gets \text{Softmax}(W_r \cdot R + b_r)$
\State $r_{pred} \gets \arg\max P(R)$
\State \Return $r_{pred}$
\end{algorithmic}
\end{algorithm}

\begin{figure*}[ht!]
	\centering 
	\includegraphics[width=0.9\textwidth]{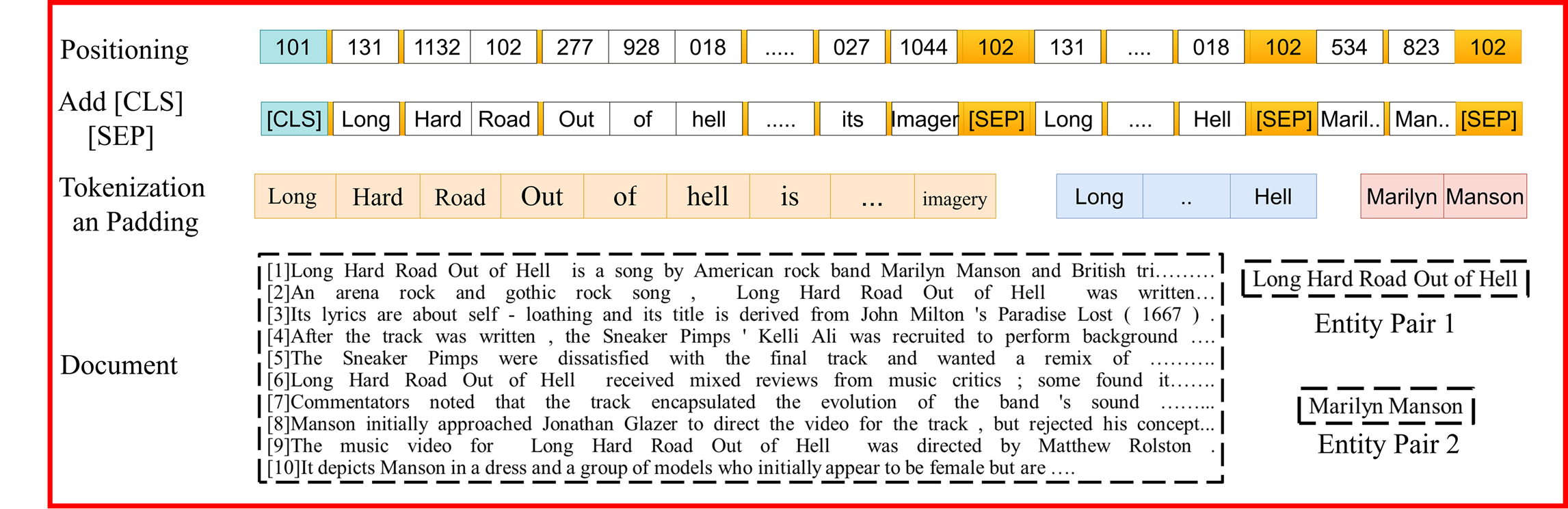}	
	\caption{Input Encoding} 
	\label{Input Encoding}
\end{figure*}

\subsection{Input Encoding}
The proposed method employs embedding tokens along with their positional and segment information to capture a document's global context and semantics as depicted in Figure \ref{Input Encoding}. It combines embeddings for document tokens, entities, and special markers like [CLS] and [SEP] into a single sequence to explicitly highlight entity pairs and their relationships within the document context. By leveraging this comprehensive representation, the model can better understand the significance of entity pairs across multiple sentences, making it particularly effective for tasks requiring deep contextual understanding. To understand mathematically, Let the document \( D \) consist of \( n \) tokens:
\begin{equation}
D = \{ t_1, t_2, \ldots, t_n \}
\end{equation}

Let \( E_1 \) represent the tokens of first entity and \( E_2 \) represent the tokens of second entity:
\begin{equation}
E_1 = \{ e_1^{(1)}, e_2^{(1)}, \ldots, e_{m_1}^{(1)} \}, \quad
E_2 = \{ e_1^{(2)}, e_2^{(2)}, \ldots, e_{m_2}^{(2)} \}
\end{equation}
Here, \( m_1 \) and \( m_2 \) are the lengths of \( E_1 \) and \( E_2 \), respectively.The input sequence includes special tokens \texttt{[CLS]} and \texttt{[SEP]}.Let the embedding of each token \( t_i \), \( e_j^{(1)} \), and \( e_j^{(2)} \) be denoted by \( \text{Embed}(x) \). Each token in the sequence is assigned a positional embedding \( \text{Pos}(i) \), indicating its position in the sequence. The document tokens \( D \) are assigned a segment embedding \( \text{Seg}_A \).  
The entity texts \( E_1 \) and \( E_2 \) are assigned a segment embedding \( \text{Seg}_B \).
The input sequence is represented as a concatenation of embeddings:
\begin{dmath}
\text{X} = 
\big[ \text{Embed}(\texttt{[CLS]}) + \text{Pos}(1) + \text{Seg}_A \big] 
\oplus 
\sum_{i=1}^n \big[ \text{Embed}(t_i) + \text{Pos}(i+1) +  \text{Seg}_A \big] 
\oplus 
\big[ \text{Embed}(\texttt{[SEP]}) + \text{Pos}(n+2) + \text{Seg}_B \big] 
\oplus 
\sum_{j=1}^{m_1} \big[ \text{Embed}(e_j^{(1)}) +  \text{Pos}(n+2+j) +  \text{Seg}_B \big] 
\oplus 
\big[ \text{Embed}(\texttt{[SEP]}) + \text{Pos}(n+2+m_1+1) +  \text{Seg}_B \big] 
\oplus 
\sum_{k=1}^{m_2} \big[ \text{Embed}(e_k^{(2)}) + \text{Pos}(n+2+m_1+k+1) + \text{Seg}_B \big] 
\oplus 
\big[ \text{Embed}(\texttt{[SEP]}) + \text{Pos}(n+2+m_1+m_2+2) + \text{Seg}_B \big]
\end{dmath}
Here, \( \oplus \) denotes concatenation.  Figure \ref{Input Encoding} illustrates an abstract view of the input encoding, which treats the document context holistically and assigns explicit, separate segments for entity pairs. This approach underscores how the broader document context informs the relationship among the specified entities. Such explicit entity representation is particularly beneficial when the same entity pairs are mentioned multiple times throughout the document or when the relevant context is distributed across several sentences.

\subsection{BERT Encoder layers}

The proposed method employs a pre-trained BERT (bert-base-uncased) without any additional refinement of its encoder, thereby preserving the original BERT weights and its inherent generalization capability. Rather than extracting the complete set of hidden states, the proposed method relies solely on the pooled output, specifically, the [CLS] token following a tanh activation on the first hidden state and omits the sequence output. Instead of extracting the complete set of hidden states, the proposed encoder utilizes only the pooled output, i.e., the [CLS] token after a tanh activation on the first hidden state, while excluding the sequence output. Focusing only on this pooled representation makes the architecture simpler and more computationally efficient while capturing valuable contextual cues. Additionally, a dropout layer with a 0.3 probability is applied to the pooled output to mitigate overfitting and reduce reliance on specific neurons, further enhancing the methods’s ability to generalize to new data.

\subsection{Classification Layer}
\begin{figure*}[ht!]
	\centering 
	\includegraphics[width=0.9\textwidth]{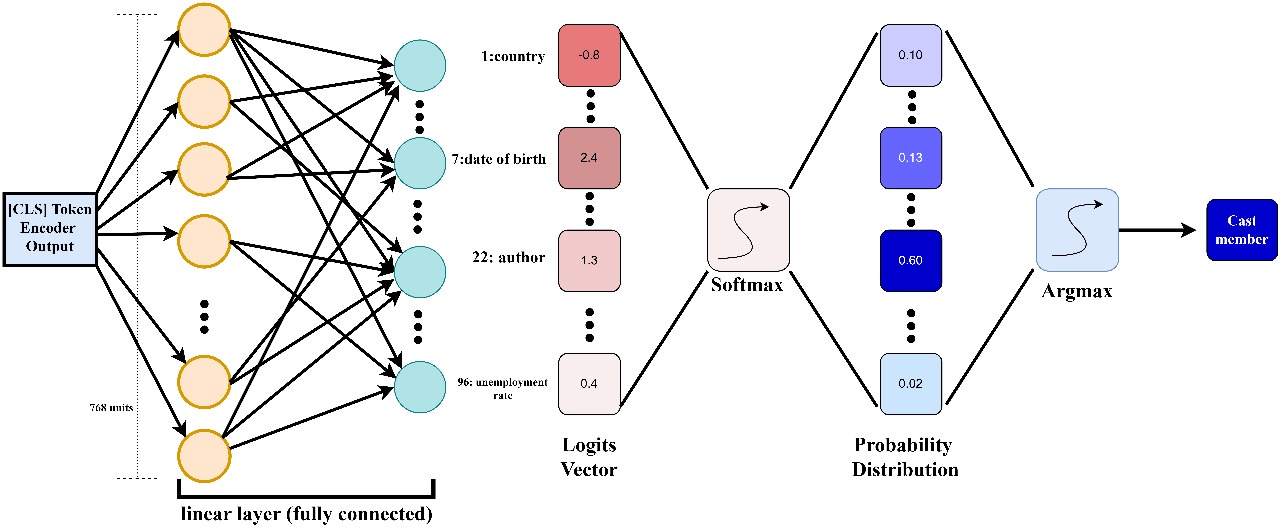}	
	\caption{Relation Classifier} 
	\label{Relation Classifier}
\end{figure*}

The classification layer utilizes the [CLS] token from BERT’s encoder to identify the appropriate class for a given pair of entities based on contextual information extracted from the document, as illustrated in Figure \ref{Relation Classifier}. The final state of the [CLS] token, which contains summarized information, is passed into a fully connected layer that maps it to a set of raw scores (logits) for each possible class. These logits represent unnormalized prediction scores, where higher values indicate a greater likelihood of the corresponding class. The logits are then fed into a softmax function, which converts them into probabilities. Finally, the class with the highest probability is selected as the final prediction.

\section{Experiments}
The performance of the proposed method has been evaluated on three benchmark datasets. The following sections present detailed information on the considered datasets, experiment settings, evaluation metrics, and results.

\subsection{Dataset}
The training and evaluation of the proposed method are done on three benchmark document-level datasets, namely DocRED, ReDocRED, and REBEL. The detailed statistics of the datasets are tabulated in Table \ref{table:dataset_statistics}.

\begin{table*}[]
\centering
\caption{Dataset Statistics}
\label{table:dataset_statistics}
\begin{tabular}{llllllll}
Dataset                       & \#Doc. & \#Word  & \#Sent. & \#Ent.  & \#Rel. & \#Inst. & \#Fact \\
DocRED (Human-annotated)      & 5053   & 1,002k  & 40276   & 132375  & 96     & 63427   & 56354  \\
DocRED (Distantly Supervised) & 101873 & 21,368k & 828115  & 2558350 & 96     & 1508320 & 881298 \\
Re-DocRED                     & 5053   & 1.5M    & 92000   & 132000  & 96     & 56000   & 240000 \\
REBEL                         & 3000   & 800k    & 45000   & 50000   & 200    & 30000   & 120000
\end{tabular}
\end{table*}

\textit{\textbf{DocRED}}: The DocRED dataset comprises data from a variety of named entity types, such as persons (18.5\%), locations (30.9\%), organizations (14.4\%), duration (15.8\%), numerals (5.1\%), and various entities like events, artistic works, and laws (15.2\%) \cite{yao2019docred}. It comprises 3,053 train_annotated, 101,873 train_distant, 998 validation, and 1,000 test instances. On average, each entity is mentioned 1.34 times. The dataset comprises 96 frequent relation types from Wikidata, organized in classes such as science (33.3\%), art (11.5\%), duration (8.3\%), and personal life (4.2\%), with a well-defined hierarchy and taxonomy. A manual analysis of 300 sampled documents revealed that 38.9\% of relation instances could be extracted through simple pattern recognition, while 61.1\% required reasoning, including logical (26.6\%), coreference (17.6\%), and commonsense (16.6\%) reasoning. Each relation instance is associated with 1.6 supporting sentences, with 46.4\% of instances linked to multiple sentences. Overall, 40.7\% of relational facts span multiple sentences, necessitating reading, synthesizing, and reasoning skills over multiple sentences.

\textit{\textbf{Re-DocRED}}: The Re-DocRED dataset is a revised version of the DocRED dataset, addressing several limitations of the original \cite{tan2022revisiting}. It incorporates a more extensive set of relation triples to tackle the problem of incompleteness, making it more comprehensive. The dataset is split into train, dev, and test sets with 3,053, 500, and 500 instances, respectively. The dataset has corrected coreferential errors and resolved logical inconsistencies present in DocRED. Like DocRED, it includes various named entity types, such as persons, locations, organizations, and miscellaneous entities like events and artistic works. Re-DocRED maintains the complexity of document-level relation extraction by covering 96 frequent relation types from Wikidata, organized into various categories such as science, art, time, and many more. The dataset emphasizes the importance of reasoning skills, including logical, coreference, and commonsense reasoning. Each relation instance in Re-DocRED is linked to multiple supporting sentences, requiring advanced reading, synthesizing, and reasoning over multiple sentences to extract relational facts accurately.

\textit{\textbf{REBEL}}: The REBEL\cite{gao2025rebel} dataset is designed for relation extraction using a seq2seq model based on BART \cite{lewis2019bart}, which simplifies the task by framing it as an end-to-end language generation problem. It covers over 200 different relation types and is trained on a large corpus of English text. The dataset includes relation triplets extracted from raw text, enabling applications such as populating knowledge bases and fact-checking. The dataset emphasizes the importance of handling multiple relation types and the flexibility of the seq2seq approach for relation extraction tasks.

\subsection{Experiment Settings and Evaluation Metrics}

\begin{figure}
	\centering 
	\includegraphics[width=0.45\textwidth]{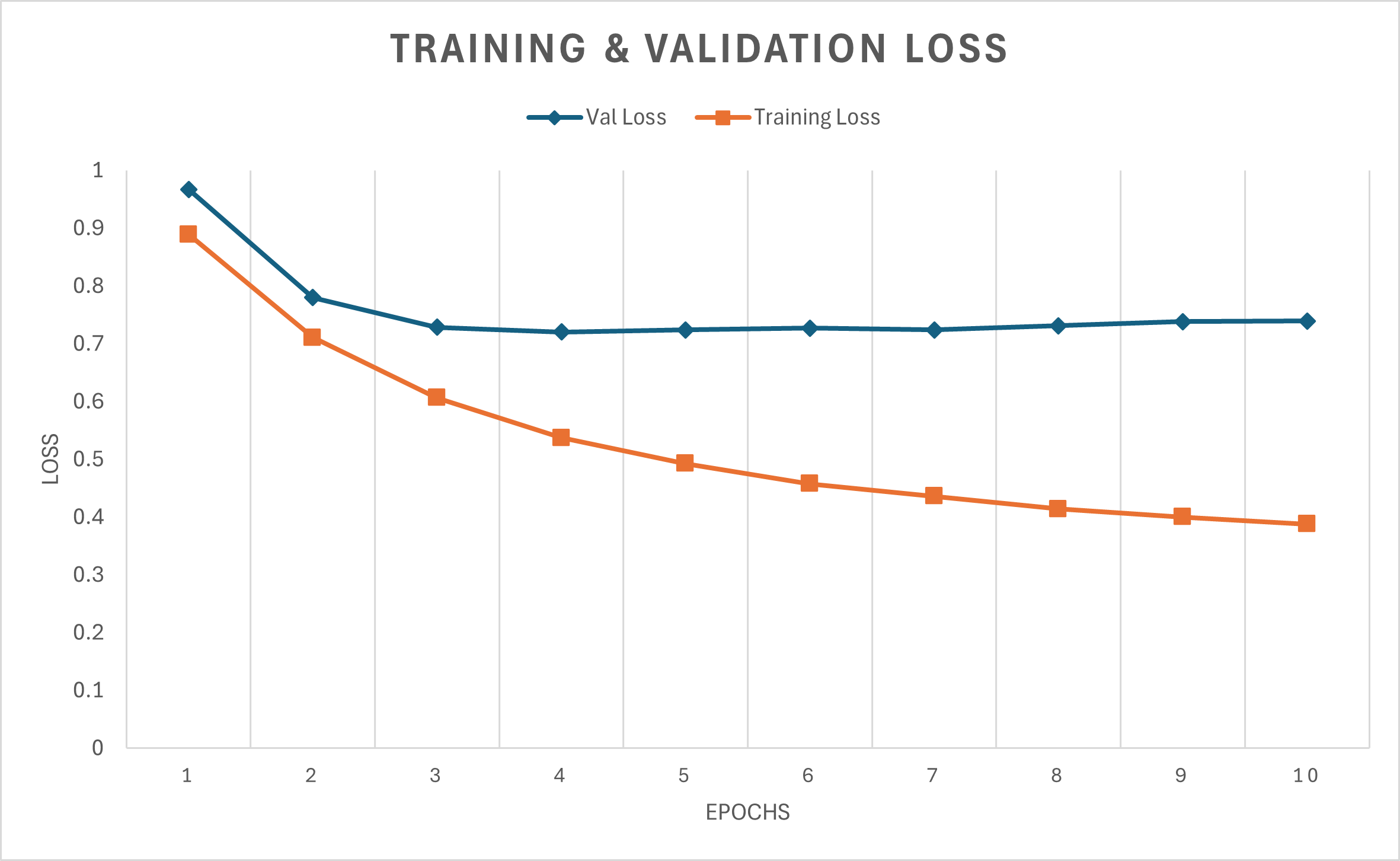}	
	\caption{Training and Validation Loss-10 Epochs} 
	\label{Training and Validation Loss-10 Epochs}
\end{figure}

To assess the performance of the proposed method, it was initially trained on the designated training dataset. In the experimental setup, 80\% of the data was used for training, while the remaining part was reserved for testing and validation. Various batch sizes, including 16, 32, and 64, were explored to evaluate the method’s effectiveness. The proposed method was also examined on the validation set for 10, 7, 4, and 3 epochs as depicted in Figure \ref{Training and Validation Loss-10 Epochs} to identify the optimal number of epochs that avoids both underfitting and overfitting. The figure shows that the loss and accuracy degrade after the third epoch, indicating that three epochs offer the best balance. The performance of the proposed method was evaluated using precision, recall, accuracy, and loss. These computations were performed on NVIDIA’s A100 SXM4 40 GB computing server using PyTorch 1.3. Besides this, a deep learning framework and necessary libraries were installed to set the environment.

\begin{table*}[]
\scriptsize
\caption{Train \& Val: DocRED\_train\_distant \quad Test: DocRED (Dev set)}
\label{Outcomes of the Proposed method on DocRED}
\begin{tabular}{lcccccccccc}
\textbf{Epoch} & \textbf{P (Val)} & \textbf{R (Val)} & \textbf{F1 (Val)} & \textbf{Loss (Val)} & \textbf{Acc (Val)} & \textbf{P (Test)} & \textbf{R (Test)} & \textbf{F1 (Test)} & \textbf{Loss (Train)} & \textbf{Acc (Train)} \\
1              & 82              & 83              & 81                & 40.50               & 83.4               & -                & -                & -                 & 57.3                 & 79.7                  \\
2              & 85              & 83              & 84                & 35.9                & 84.4               & 92.4             & 86.7             & 88.9              & 36.7                 & 84.7                  \\
3              & 88              & 86              & 87                & 33.8                & 84.7               & -                & -                & -                 & 29.6                 & 86.7                 
\end{tabular}
\end{table*}

\begin{table*}[]
\scriptsize
\caption{Train \& Val: DocRED\_train\_distant \quad Test: Re-DocRED (Test set)}
\label{Outcomes of the Proposed method on Re-DocRED}
\begin{tabular}{lcccccccccc}
\textbf{Epoch} & \textbf{P (Val)} & \textbf{R (Val)} & \textbf{F1 (Val)} & \textbf{Loss (Val)} & \textbf{Acc (Val)} & \textbf{P (Test)} & \textbf{R (Test)} & \textbf{F1 (Test)} & \textbf{Loss (Train)} & \textbf{Acc (Train)} \\
1              & 82               & 83               & 81                & 40.50               & 83.4               & -                 & -                 & -                  & 57.3                  & 79.7                  \\
2              & 85               & 83               & 84                & 35.9                & 84.4               & 74.6              & 67.6              & 67.2               & 36.7                  & 84.7                  \\
3              & 88               & 86               & 87                & 33.8                & 84.7               & -                 & -                 & -                  & 29.6                  & 86.7                 
\end{tabular}
\end{table*}

\subsection{Results}

To evaluate the performance of the proposed method, precision, recall, F1 scores, loss, and accuracy were computed for both the validation and test sets of the DocRED, Re-DocRED, and REBEL datasets. The DocRED and Re-DocRED datasets have been used in various NLP tasks, including named entity recognition, topic labeling, and document-level relation extraction. On the contrary, the REBEL dataset is generally used for sentence-level relation extraction. Therefore, to affirm the efficacy of the proposed method, it has been validated on both document-level and sentence-level datasets. Table \ref{Outcomes of the Proposed method on DocRED} presents the outcomes of the proposed method on DocRED, while Tables \ref{Outcomes of the Proposed method on Re-DocRED} and \ref{Outcomes of the Proposed method on REBEL} summarize the performance on Re-DocRED and REBEL datasets, respectively. It is evident from the tables that the mean values of all performance metrics on both validation and test sets are nearly identical, indicating the proposed method's strong generalization capability on unseen data.

\begin{table*}[]
\scriptsize
\caption{Train \& Val: REBEL's Training and Validation Set, \quad Test: REBEL (Test Set)}
\label{Outcomes of the Proposed method on REBEL}
\begin{tabular}{lcccccccccc}
\textbf{Epoch} & \textbf{P (Val)} & \textbf{R (Val)} & \textbf{F1 (Val)} & \textbf{Loss (Val)} & \textbf{Acc (Val)} & \textbf{P (Test)} & \textbf{R (Test)} & \textbf{F1 (Test)} & \textbf{Loss (Train)} & \textbf{Acc (Train)} \\
1              & 92.2             & 92.8             & 92.3              & 29.3                & 93.0               & 93.5              & 93.9              & 93.6               & 41.7                  & 90.8                  \\
2              & 93.3             & 93.6             & 93.3              & 27.8                & 94.0               & -                 & -                 & -                  & 21.6                  & 94.7                  \\
3              & 93.8             & 94.0             & 93.8              & 28.8                & 94.0               & -                 & -                 & -                  & 14.5                  & 96.5                 
\end{tabular}
\end{table*}

Furthermore, the proposed method has been validated against state-of-the-art methods, including, SIRE \cite{zeng2021sire}, DRN \cite{xu2021discriminative},, HeterGSAN-Rec \cite{xu2021document},GAIN \cite{zeng2020double}, ATLOP \cite{zhou2021document}, DocuNet \cite{zhang2021document}, SSR-PU \cite{wang2022unified}, KD-DocRE \cite{tan2022document}, SAIS \cite{xiao2021sais}, DREEAM \cite{ma2023dreeam}, EIDER \cite{xie2021eider}, KIRE \cite{wang2022enhancing}, RESIDE \cite{vashishth2018reside}, RECON \cite{bastos2021recon}, KB-Graph \cite{verlinden2021injecting}, and DocRE-CLiP \cite{jain2024revisiting} in terms of F1 score. GAIN \cite{zeng2020double}, KIRE \cite{wang2022enhancing}, and KB-Graph \cite{verlinden2021injecting} utilize graphs to reason about relationships between entity pairs, while ATLOP \cite{zhou2021document}, SSR-PU \cite{wang2022unified}, and KD-DocRE \cite{tan2022document} employ transformer learning frameworks and distillation techniques for relation extraction. The other remaining methods, such as DocuNet \cite{zhang2021document}, SAIS \cite{xiao2021sais}, etc., use semantic information and contrastive learning for relation extraction between any pair of entities. Table \ref{Result compared of proposed method to DocRED dataset} presents the results of the proposed and state-of-the-art methods on the DocRED dataset. The bold fonts in the table represent the best values. It can be observed from Table \ref{Result compared of proposed method to DocRED dataset} that the proposed method attained 84\% F1 score on the validation set and 88.91\% on the test set of the DocRed dataset. In contrast, the top-performing existing method, DocRE-CLiP \cite{jain2024revisiting}, yielded validation and test F1 scores of 66.43\% and 66.31\%, respectively.  Hence, Table \ref{Result compared of proposed method to DocRED dataset} demonstrates the effectiveness of the proposed method on DocRED. A similar pattern emerges in Table \ref{Result compared of proposed method to Re-DocRED dataset} for the Re-DocRED dataset. The proposed method attained an 84\% of F1 score on the validation set and 67.19\% on the test set, surpassing other approaches by a significant margin. The best-performing existing method achieved 67.41\% on the validation set and 67.53\% on the test set, indicating a 24.61\% increase on the validation set and comparable performance on the test set. 
Overall, the above findings suggest that the proposed method consistently delivers superior results across all evaluation metrics, reinforcing its advantages over existing methods.

\begin{table*}[]
\centering
\scriptsize
\caption{Comparison of the proposed method on the DocRED Dataset}
\label{Result compared of proposed method to DocRED dataset}
\begin{tabular}{lcc|cc}
\textbf{Model} & \textbf{PLM/GNN} &  & \textbf{Dev F1 (\%)} & \textbf{Test F1 (\%)} \\
SIRE (2021)            & BERT     &   & 59.82 & 60.18 \\
HeterGSAN-Rec (2021)   & BERT     &   & 58.13 & 57.12 \\
ATLOP (2021)           & BERT     &   & 59.22 & 59.31 \\
DRN (2021)             & BERT     &   & 59.33 & 59.15 \\
DocuNet (2021)         & RoBERTa  &   & 62.23 & 62.39 \\
ATLOP (2021)           & RoBERTa  &   & 61.32 & 61.39 \\
KD-DocRE (2022)        & RoBERTa  &   & 65.27 & 65.24 \\
SAIS (2021)            & RoBERTa  &   & 62.23 & 63.44 \\
DREEAM (2024)          & RoBERTa  &   & 65.52 & 65.47 \\
EIDER (2021)           & RoBERTa  &   & 62.34 & 62.85 \\
KIRE (2022)            & --       &   & 50.46 & 49.69 \\
RESIDE (2018)          & GNN      &   & 49.64 & 48.62 \\
RECON (2021)           & GNN      &   & 50.78 & 49.97 \\
KB-Graph (2021)        & --       &   & 50.69 & 49.88 \\
DocRE-CLiP (2024)      & BERT     &   & 66.43 & 66.31 \\
\textbf{Proposed Method} & \textbf{BERT} &   & \textbf{84.00} & \textbf{88.91} \\
\end{tabular}
\end{table*}

\begin{table*}[]
\centering
\scriptsize
\caption{Comparison of the proposed method on the Re-DocRED Dataset}
\label{Result compared of proposed method to Re-DocRED dataset}
\begin{tabular}{p{4.4cm}llc}
\textbf{Model} & \textbf{PLM} & \textbf{Val F1} & \textbf{Test F1} \\
ATLOP (2021)                          & BERT     & -                 & 43.25 ± 0.25 \\
GAIN (2020)                           & BERT     & -                 & 45.82 ± 1.38 \\
KD-DocRE (2022)                       & BERT     & 67.12 ± 0.14      & 67.28       \\
DocuNet (2021)                        & RoBERTa  & -                 & 45.99 ± 0.33 \\
DREEAM (2024)                         & RoBERTa  & 67.41 ± 0.04      & 67.53       \\
SSR-PU+ATLOP+ BERT (2022)             & BERT     & -                 & 56.14 ± 0.12 \\
SSR-PU+ATLOP+ RoBERTa (2022)          & RoBERTa  & -                 & 59.50 ± 0.45 \\
\textbf{Proposed Method}             & \textbf{BERT} & \textbf{-}     & \textbf{67.19} \\
\end{tabular}
\end{table*}

\subsubsection{State-of-the-Art (SOTA) Vs. EPGICo}
Despite the significant advancements in document-level relation extraction (DocRE), state-of-the-art (SOTA) models still exhibit notable limitations, particularly in capturing global context and effectively handling long-range dependencies. For instance, ATLOP (2021) incorporates adaptive thresholding for overlapping relations but lacks mechanisms for aggregating explicit global context, while GAIN (2020), a graph-based approach, models inter-sentence relationships yet struggles to fully leverage recurrent entity information. Similarly, KD-DocRE (2022) improves training efficiency through knowledge distillation but at the expense of losing finer contextual nuances. Models such as DocuNet (2021) and DREEAM (2024) face challenges with sequential dependencies and complex multi-hop reasoning, respectively. SSR-PU+ATLOP addresses selective self-relation mechanisms but lacks a structured global context representation. Moreover, HeterGSAN-Rec (2021) relies on heterogeneous graph attention but does not adequately model entity mentions across entire documents, and DocRE-CLiP (2024), though utilizing contrastive learning, remains constrained by sentence-level information. These approaches perform well in specific scenarios but fail to leverage multiple occurrences of entities in different contexts throughout a document, resulting in incomplete or suboptimal relation extraction outcomes.

The input processing in state-of-the-art (SOTA) models for document-level relation extraction typically involves encoding the entire document and entity information. For instance, models like ATLOP rely on adaptive thresholding but often treat entity pairs in isolation without explicitly aggregating global context. Graph-based models like GAIN construct entity graphs to capture inter-sentence dependencies but may struggle with long-range relationships across the document. Due to simplified teacher-student mechanisms, KD-DocRE leverages knowledge distillation but can miss finer contextual nuances. In contrast, the proposed model incorporates a unified global context mechanism by concatenating the document and entities into a single input sequence, enabling all occurrences of head and tail entities to be explicitly modeled. This input representation is further enriched with positional and type embeddings, ensuring a deeper contextual understanding. The proposed model effectively aggregates global context by leveraging a 12-layer Transformer encoder and multi-head self-attention. This leads to more accurate relation extraction than SOTA models, which often fail to comprehensively handle entity recurrence and long-range dependencies.
\subsubsection{Analysis of Proposed Method}
Furthermore, Figure \ref{Precision, Recall, and F1 Score Curve} shows the proposed method's precision, recall, and F1 score on the DocRED dataset. As illustrated, precision increases from 81\% in the first epoch to 83\% in the second epoch, after which it stabilizes. Likewise, recall and F1 scores improve over three epochs, rising from 78\% to 81\% and from 79\% to 82\%, respectively. The consistent rise in recall and F1 score implies that the proposed method is more adept at identifying relevant cases while balancing precision and recall over time. In addition, the precision stability at 0.83 after the second epoch shows that the proposed method predicts positive instances more accurately, even as recall improves. 

\begin{figure}[ht]
	\centering 
	\includegraphics[width=0.5\textwidth]{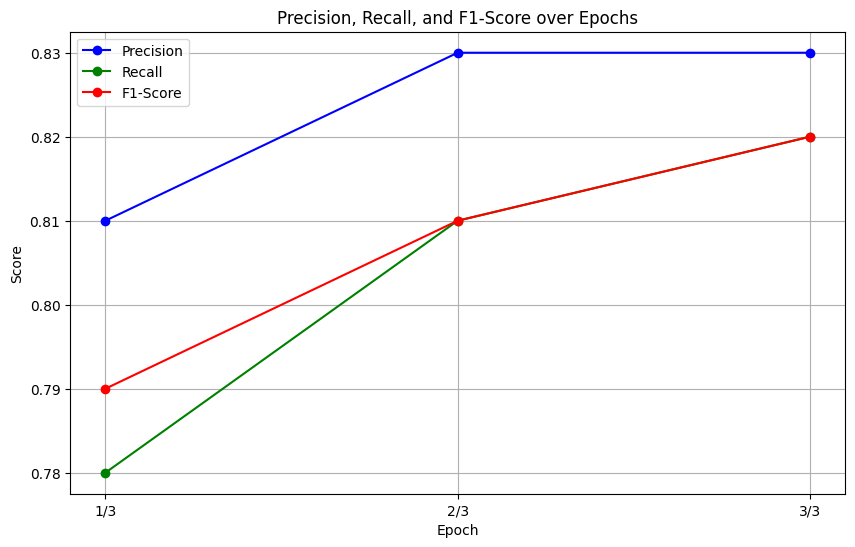}	
	\caption{Precision, Recall, and F1 Score Curve} 
	\label{Precision, Recall, and F1 Score Curve}
\end{figure}
\begin{figure}[ht]
	\centering 
	\includegraphics[width=0.5\textwidth]{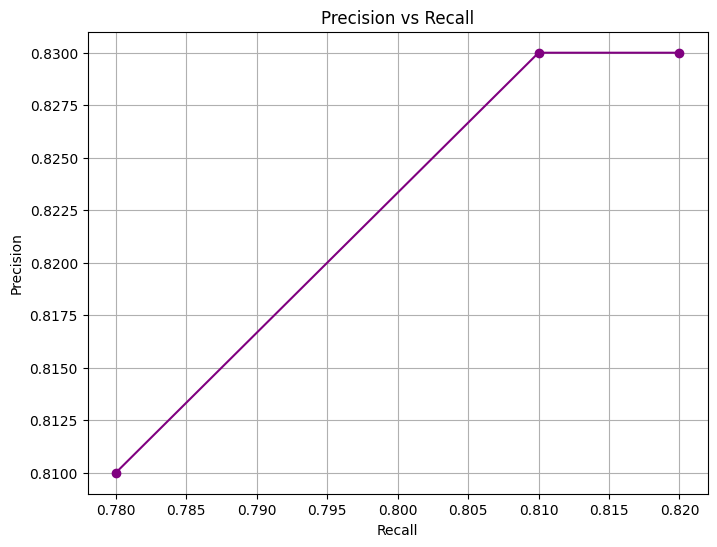}	
	\caption{Precision and Recall Curve} 
	\label{Precision and Recall Curve}
\end{figure}

\begin{figure*}[htbp!]
	\centering 
	\includegraphics[width=0.8\textwidth]{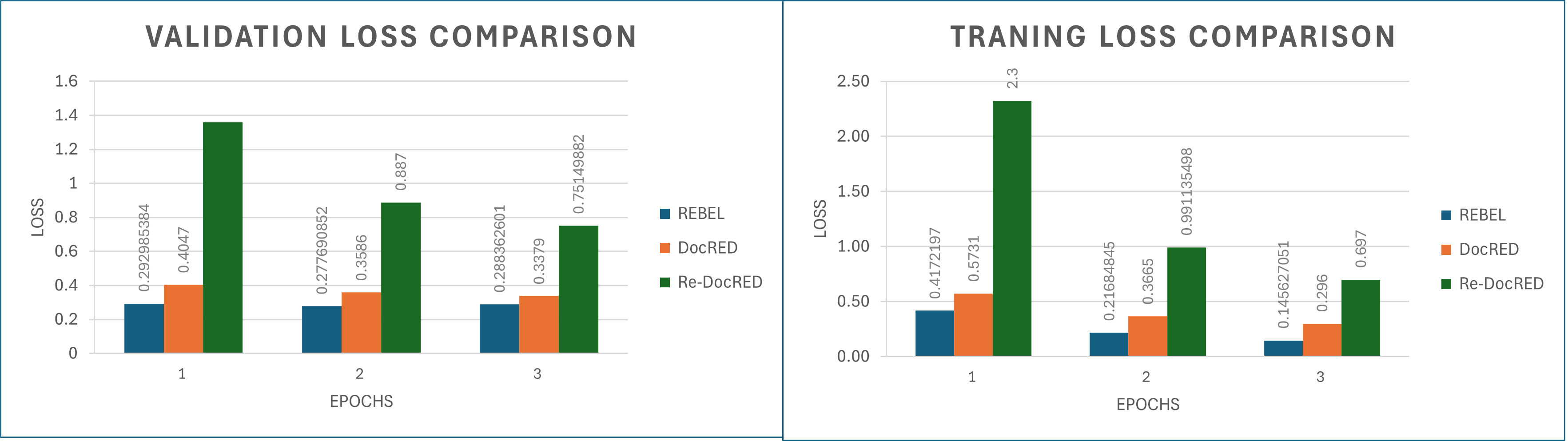}	
	\caption{Loss Comparision of evaluation matrics on three Dataset} 
	\label{Loss Comparision of evaluation matrics on three Dataset}
\end{figure*}

\begin{figure*}[htbp!]
	\centering 
	\includegraphics[width=0.8\textwidth]{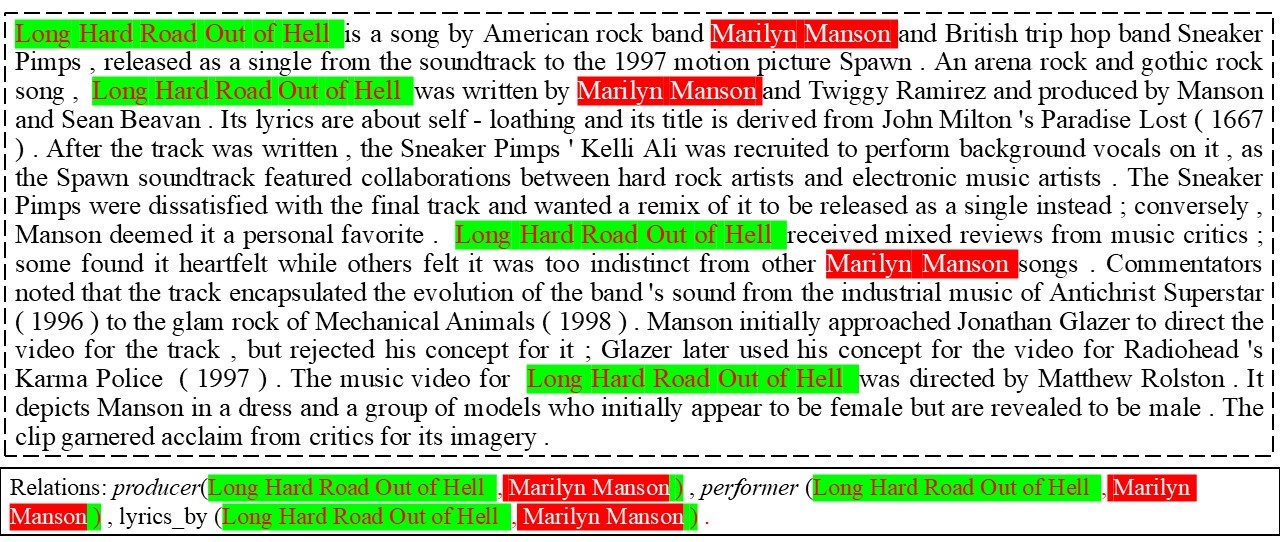}	
	\caption{Example of DRE. Note that mentions of the same entity are marked with identical colors. The output of the proposed method on the above example; performer(Long Hard Road Out of Hell, Marilyn Manson)} 
	\label{Local-Global Relation1}
\end{figure*}

\begin{figure*}[htbp!]
	\centering 
	\includegraphics[width=0.8\textwidth]{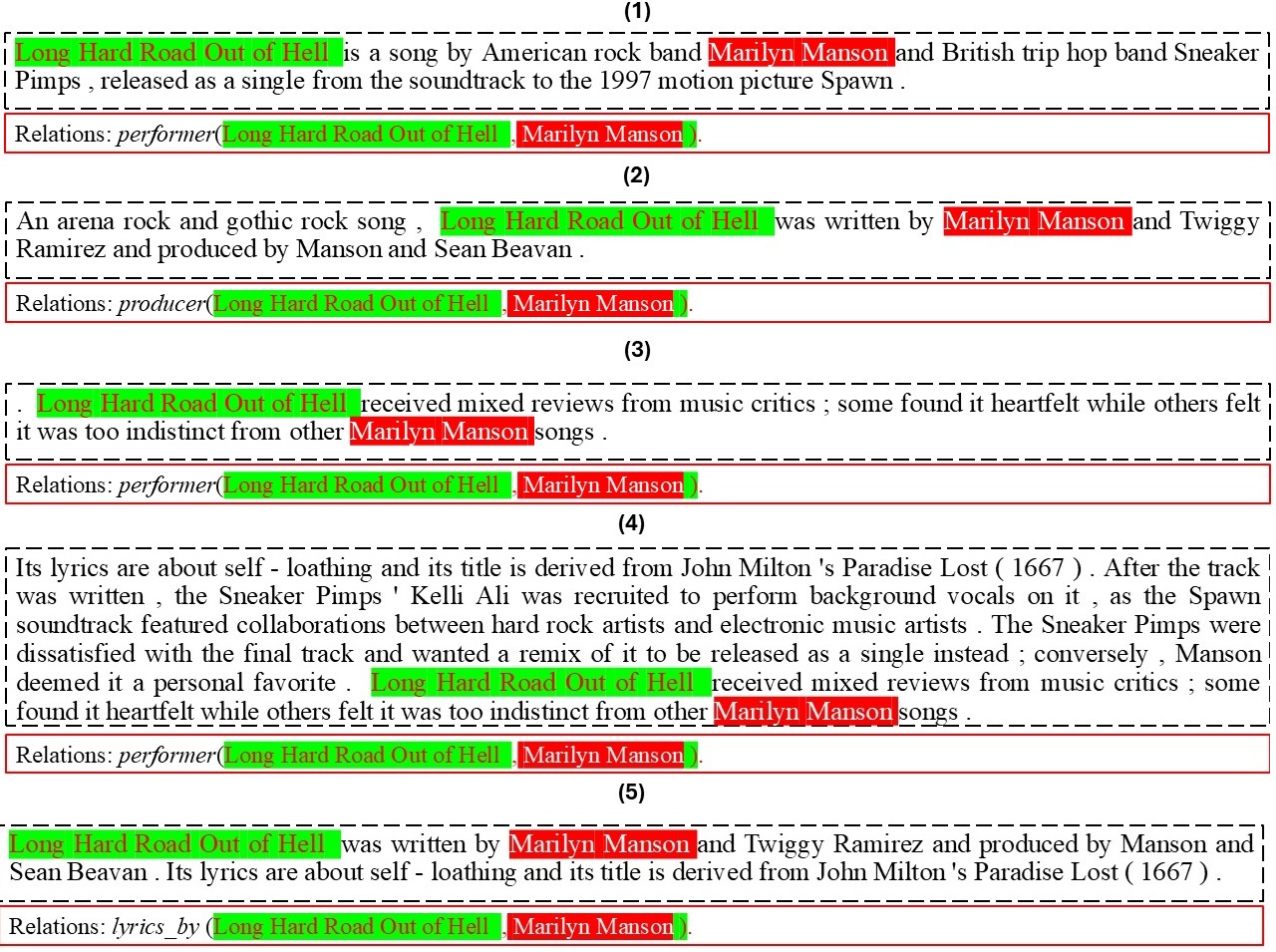}	
	\caption{Output Relations of the proposed method on same entity pair on different Local context} 
	\label{Local-Global Relation2}
\end{figure*}

Moreover, a precision-recall curve is also drawn in Figure \ref{Precision and Recall Curve} for the DocRed dataset to highlight the proposed method’s overall effectiveness. It can be observed from the figure that the proposed method demonstrates strong performance in identifying true positives while maintaining a high rate of accurate predictions. An increase in precision and recall indicates that the proposed method is becoming more accurate (higher precision) and detecting a more significant number of true positives (higher recall). Besides, Figure \ref{Loss Comparision of evaluation matrics on three Dataset} illustrates the training and validation loss of the proposed method over three epochs of all the considered datasets. It is evident from the figure that the training and validation loss of the proposed method on the REBEl dataset decreased from 0.417 to 0.145 and 0.292 to 0.288 over three epochs, respectively, indicating intense learning and generalization capabilities of the proposed method on the REBEl dataset. On the contrary, the training and validation loss of the proposed method on the DocRed dataset over three epochs decreased from 0.573 to 0.296 and 0.404 to 0.337, respectively. From the outcomes, it can be envisioned that the proposed method also performs well on the DocRed dataset. However, its slightly higher loss indicates a marginal reduction in efficiency compared to REBEL due to more false negative instances. In contrast, Re-DocRED experiences a significant drop in training and validation loss after the first epoch from 2.3 to 0.697 and 1.35 to 0.751, which indicates that, although it initially struggles, it can improve, given additional training epochs. Thus, with more epochs, Re-DocRED may narrow the gap and achieve comparable stability and performance. From the above analysis, it has been observed that the proposed method showed promising potential on all the considered datasets. However, reducing false negative instances in the DocRED and Re-DocRED datasets remains crucial for further performance improvements. 

Further, to affirm the efficacy of the proposed method, a different set of contexts has been given for the same pair of entities to predict the correct relationships at both the local and global levels of a document's context. For example, the entity pair ``Long Hard Road Out of Hell" and ``Marilyn Manson" appear multiple times in the document, and each occurrence is associated with a different relationship depending on the local context, as shown in Figure \ref{Local-Global Relation1}. The proposed method predicts the appropriate relationship between entity pair ``Long Hard Road Out of Hell" and ``Marilyn Manson" based on the provided context as in Figure \ref{Local-Global Relation2}. The analysis confirms that the proposed method effectively captures contextual nuances and accurately determines entity pair relationships.

\begin{table*}[]
\centering
\caption{Results of the proposed method}
\label{Results of the proposed method}
\begin{tabular}{lcccccc}
\textbf{Dataset} & \textbf{F1 (Dev)} & \textbf{P (Dev)} & \textbf{R (Dev)} & \textbf{F1 (Test)} & \textbf{P (Test)} & \textbf{R (Test)} \\
DocRED           & 84.12             & 85.21            & 84.01            & 88.91              & 92.39             & 86.73             \\
Re-DocRED        & -                 & -                & -                & 67.19              & 74.61             & 67.58             \\
REBEL            & 93.14             & 93.07            & 93.48            & 93.59              & 93.48             & 93.88             
\end{tabular}
\end{table*}

\section{Conclusions}
This article presented a BERT-based model for document-level relationship extraction. The proposed method incorporated the complete context of the document and entity pair to improve DRE. 
Rather than focusing solely on the local context around each entity, the proposed method examines the global context to identify the most appropriate relation, even when evidence is distributed across the text. Rather than finding the local context of the corresponding entity, the proposed method works on the global context of the document to find the most suitable relation between the entity pair that may be spread throughout the text. An input encoding strategy is also proposed to better capture the contextual dependencies between relations and entity pairs.
Three benchmark relation extraction datasets have been used to assess the performance of the proposed method. The proposed method has also been compared with state-of-the-art methods regarding precision, recall, and F1 score. The proposed method attained a higher precision, recall, and F1 score on validation and test sets on all the considered datasets, which validated its performance. Besides, the proposed method predicts the appropriate relationship for the same pair of entities with different contexts as it interprets contextual nuances correctly.
While the proposed method demonstrated improved efficiency over existing approaches on the DocRed and REBEL datasets, its performance on the Re-DocRED dataset still requires enhancement. Due to the probabilistic nature, the model sometimes predicts relationships between entity pairs in Re-DocRED that do not exist. Therefore, future work involves incorporating a new reasoning component in the proposed method to tackle the false negative instances of the Re-DocRED dataset.

\bibliographystyle{IEEEtran}

\begin{thebibliography}{10}
\providecommand{\url}[1]{#1}
\csname url@samestyle\endcsname
\providecommand{\newblock}{\relax}
\providecommand{\bibinfo}[2]{#2}
\providecommand{\BIBentrySTDinterwordspacing}{\spaceskip=0pt\relax}
\providecommand{\BIBentryALTinterwordstretchfactor}{4}
\providecommand{\BIBentryALTinterwordspacing}{\spaceskip=\fontdimen2\font plus
\BIBentryALTinterwordstretchfactor\fontdimen3\font minus \fontdimen4\font\relax}
\providecommand{\BIBforeignlanguage}[2]{{%
\expandafter\ifx\csname l@#1\endcsname\relax
\typeout{** WARNING: IEEEtran.bst: No hyphenation pattern has been}%
\typeout{** loaded for the language `#1'. Using the pattern for}%
\typeout{** the default language instead.}%
\else
\language=\csname l@#1\endcsname
\fi
#2}}
\providecommand{\BIBdecl}{\relax}
\BIBdecl

\bibitem{pawar2017relation}
S.~Pawar, G.~K. Palshikar, and P.~Bhattacharyya, ``Relation extraction: A survey,'' \emph{arXiv preprint arXiv:1712.05191}, 2017.

\bibitem{delaunay2023comprehensive}
J.~Delaunay, H.~T.~H. Tran, C.-E. Gonz{\'a}lez-Gallardo, G.~Bordea, N.~Sidere, and A.~Doucet, ``A comprehensive survey of document-level relation extraction (2016-2023),'' \emph{arXiv preprint arXiv:2309.16396}, 2023.

\bibitem{zhao2024comprehensive}
X.~Zhao, Y.~Deng, M.~Yang, L.~Wang, R.~Zhang, H.~Cheng, W.~Lam, Y.~Shen, and R.~Xu, ``A comprehensive survey on relation extraction: Recent advances and new frontiers,'' \emph{ACM Computing Surveys}, vol.~56, no.~11, pp. 1--39, 2024.

\bibitem{sun2022dual}
Q.~Sun, T.~Xu, K.~Zhang, K.~Huang, L.~Lv, X.~Li, T.~Zhang, and D.~Dore-Natteh, ``Dual-channel and hierarchical graph convolutional networks for document-level relation extraction,'' \emph{Expert Systems with Applications}, vol. 205, p. 117678, 2022.

\bibitem{jia2024document}
W.~Jia, R.~Ma, L.~Yan, W.~Niu, and Z.~Ma, ``Document-level relation extraction with global and path dependencies,'' \emph{Knowledge-Based Systems}, vol. 289, p. 111545, 2024.

\bibitem{zhao2024corex}
S.~Zhao and C.~Li, ``Corex: Document-level relation extraction framework with consistent two-hop reasoning and evidence sentence prediction,'' in \emph{2024 International Joint Conference on Neural Networks (IJCNN)}.\hskip 1em plus 0.5em minus 0.4em\relax IEEE, 2024, pp. 1--8.

\bibitem{sahu2019intersentencerelationextractiondocumentlevel}
\BIBentryALTinterwordspacing
S.~K. Sahu, F.~Christopoulou, M.~Miwa, and S.~Ananiadou, ``Inter-sentence relation extraction with document-level graph convolutional neural network,'' 2019. [Online]. Available: \url{https://arxiv.org/abs/1906.04684}
\BIBentrySTDinterwordspacing

\bibitem{Zhou_Huang_Ma_Huang_2021}
\BIBentryALTinterwordspacing
W.~Zhou, K.~Huang, T.~Ma, and J.~Huang, ``Document-level relation extraction with adaptive thresholding and localized context pooling,'' \emph{Proceedings of the AAAI Conference on Artificial Intelligence}, vol.~35, no.~16, pp. 14\,612--14\,620, May 2021. [Online]. Available: \url{https://ojs.aaai.org/index.php/AAAI/article/view/17717}
\BIBentrySTDinterwordspacing

\bibitem{Xu_Wang_Lyu_Zhu_Mao_2021}
\BIBentryALTinterwordspacing
B.~Xu, Q.~Wang, Y.~Lyu, Y.~Zhu, and Z.~Mao, ``Entity structure within and throughout: Modeling mention dependencies for document-level relation extraction,'' \emph{Proceedings of the AAAI Conference on Artificial Intelligence}, vol.~35, no.~16, pp. 14\,149--14\,157, May 2021. [Online]. Available: \url{https://ojs.aaai.org/index.php/AAAI/article/view/17665}
\BIBentrySTDinterwordspacing

\bibitem{zeng2014relation}
D.~Zeng, K.~Liu, S.~Lai, G.~Zhou, and J.~Zhao, ``Relation classification via convolutional deep neural network,'' in \emph{Proceedings of COLING 2014, the 25th international conference on computational linguistics: technical papers}, 2014, pp. 2335--2344.

\bibitem{cai2016bidirectional}
R.~Cai, X.~Zhang, and H.~Wang, ``Bidirectional recurrent convolutional neural network for relation classification,'' in \emph{Proceedings of the 54th Annual Meeting of the Association for Computational Linguistics (Volume 1: Long Papers)}, 2016, pp. 756--765.

\bibitem{zheng2021prgc}
H.~Zheng, R.~Wen, X.~Chen, Y.~Yang, Y.~Zhang, Z.~Zhang, N.~Zhang, B.~Qin, M.~Xu, and Y.~Zheng, ``Prgc: Potential relation and global correspondence based joint relational triple extraction,'' \emph{arXiv preprint arXiv:2106.09895}, 2021.

\bibitem{yao2019docred}
Y.~Yao, D.~Ye, P.~Li, X.~Han, Y.~Lin, Z.~Liu, Z.~Liu, L.~Huang, J.~Zhou, and M.~Sun, ``Docred: A large-scale document-level relation extraction dataset,'' \emph{arXiv preprint arXiv:1906.06127}, 2019.

\bibitem{zhou2021document}
W.~Zhou, K.~Huang, T.~Ma, and J.~Huang, ``Document-level relation extraction with adaptive thresholding and localized context pooling,'' in \emph{Proceedings of the AAAI conference on artificial intelligence}, vol.~35, no.~16, 2021, pp. 14\,612--14\,620.

\bibitem{christopoulou2019connecting}
F.~Christopoulou, M.~Miwa, and S.~Ananiadou, ``Connecting the dots: Document-level neural relation extraction with edge-oriented graphs,'' \emph{arXiv preprint arXiv:1909.00228}, 2019.

\bibitem{quirk2016distant}
C.~Quirk and H.~Poon, ``Distant supervision for relation extraction beyond the sentence boundary,'' \emph{arXiv preprint arXiv:1609.04873}, 2016.

\bibitem{nan2020reasoning}
G.~Nan, Z.~Guo, I.~Sekuli{\'c}, and W.~Lu, ``Reasoning with latent structure refinement for document-level relation extraction,'' \emph{arXiv preprint arXiv:2005.06312}, 2020.

\bibitem{wang2020global}
D.~Wang, W.~Hu, E.~Cao, and W.~Sun, ``Global-to-local neural networks for document-level relation extraction,'' \emph{arXiv preprint arXiv:2009.10359}, 2020.

\bibitem{zeng2020double}
S.~Zeng, R.~Xu, B.~Chang, and L.~Li, ``Double graph based reasoning for document-level relation extraction,'' \emph{arXiv preprint arXiv:2009.13752}, 2020.

\bibitem{li2020graph}
B.~Li, W.~Ye, Z.~Sheng, R.~Xie, X.~Xi, and S.~Zhang, ``Graph enhanced dual attention network for document-level relation extraction,'' in \emph{Proceedings of the 28th international conference on computational linguistics}, 2020, pp. 1551--1560.

\bibitem{zhang2020document}
Z.~Zhang, B.~Yu, X.~Shu, T.~Liu, H.~Tang, W.~Yubin, and L.~Guo, ``Document-level relation extraction with dual-tier heterogeneous graph,'' in \emph{Proceedings of the 28th International Conference on Computational Linguistics}, 2020, pp. 1630--1641.

\bibitem{li2021mrn}
J.~Li, K.~Xu, F.~Li, H.~Fei, Y.~Ren, and D.~Ji, ``Mrn: A locally and globally mention-based reasoning network for document-level relation extraction,'' in \emph{Findings of the Association for Computational Linguistics: ACL-IJCNLP 2021}, 2021, pp. 1359--1370.

\bibitem{xu2021document}
W.~Xu, K.~Chen, and T.~Zhao, ``Document-level relation extraction with reconstruction,'' in \emph{Proceedings of the AAAI Conference on Artificial Intelligence}, vol.~35, no.~16, 2021, pp. 14\,167--14\,175.

\bibitem{xu2021entity}
B.~Xu, Q.~Wang, Y.~Lyu, Y.~Zhu, and Z.~Mao, ``Entity structure within and throughout: Modeling mention dependencies for document-level relation extraction,'' in \emph{Proceedings of the AAAI conference on artificial intelligence}, vol.~35, no.~16, 2021, pp. 14\,149--14\,157.

\bibitem{ru2021learning}
D.~Ru, C.~Sun, J.~Feng, L.~Qiu, H.~Zhou, W.~Zhang, Y.~Yu, and L.~Li, ``Learning logic rules for document-level relation extraction,'' \emph{arXiv preprint arXiv:2111.05407}, 2021.

\bibitem{zhang2021modular}
S.~Zhang, C.~Wong, N.~Usuyama, S.~Jain, T.~Naumann, and H.~Poon, ``Modular self-supervision for document-level relation extraction,'' \emph{arXiv preprint arXiv:2109.05362}, 2021.

\bibitem{tan2022revisiting}
Q.~Tan, L.~Xu, L.~Bing, H.~T. Ng, and S.~M. Aljunied, ``Revisiting docred--addressing the false negative problem in relation extraction,'' \emph{arXiv preprint arXiv:2205.12696}, 2022.

\bibitem{gao2025rebel}
Z.~Gao, J.~Chang, W.~Zhan, O.~Oertell, G.~Swamy, K.~Brantley, T.~Joachims, D.~Bagnell, J.~D. Lee, and W.~Sun, ``Rebel: Reinforcement learning via regressing relative rewards,'' \emph{Advances in Neural Information Processing Systems}, vol.~37, pp. 52\,354--52\,400, 2025.

\bibitem{lewis2019bart}
M.~Lewis, Y.~Liu, N.~Goyal, M.~Ghazvininejad, A.~Mohamed, O.~Levy, V.~Stoyanov, and L.~Zettlemoyer, ``Bart: Denoising sequence-to-sequence pre-training for natural language generation, translation, and comprehension,'' \emph{arXiv preprint arXiv:1910.13461}, 2019.

\bibitem{zeng2021sire}
S.~Zeng, Y.~Wu, and B.~Chang, ``Sire: Separate intra-and inter-sentential reasoning for document-level relation extraction,'' \emph{arXiv preprint arXiv:2106.01709}, 2021.

\bibitem{xu2021discriminative}
W.~Xu, K.~Chen, and T.~Zhao, ``Discriminative reasoning for document-level relation extraction,'' \emph{arXiv preprint arXiv:2106.01562}, 2021.

\bibitem{zhang2021document}
N.~Zhang, X.~Chen, X.~Xie, S.~Deng, C.~Tan, M.~Chen, F.~Huang, L.~Si, and H.~Chen, ``Document-level relation extraction as semantic segmentation,'' \emph{arXiv preprint arXiv:2106.03618}, 2021.

\bibitem{wang2022unified}
Y.~Wang, X.~Liu, W.~Hu, and T.~Zhang, ``A unified positive-unlabeled learning framework for document-level relation extraction with different levels of labeling,'' \emph{arXiv preprint arXiv:2210.08709}, 2022.

\bibitem{tan2022document}
Q.~Tan, R.~He, L.~Bing, and H.~T. Ng, ``Document-level relation extraction with adaptive focal loss and knowledge distillation,'' \emph{arXiv preprint arXiv:2203.10900}, 2022.

\bibitem{xiao2021sais}
Y.~Xiao, Z.~Zhang, Y.~Mao, C.~Yang, and J.~Han, ``Sais: supervising and augmenting intermediate steps for document-level relation extraction,'' \emph{arXiv preprint arXiv:2109.12093}, 2021.

\bibitem{ma2023dreeam}
Y.~Ma, A.~Wang, and N.~Okazaki, ``Dreeam: Guiding attention with evidence for improving document-level relation extraction,'' \emph{arXiv preprint arXiv:2302.08675}, 2023.

\bibitem{xie2021eider}
Y.~Xie, J.~Shen, S.~Li, Y.~Mao, and J.~Han, ``Eider: Empowering document-level relation extraction with efficient evidence extraction and inference-stage fusion,'' \emph{arXiv preprint arXiv:2106.08657}, 2021.

\bibitem{wang2022enhancing}
X.~Wang, Z.~Wang, W.~Sun, and W.~Hu, ``Enhancing document-level relation extraction by entity knowledge injection,'' in \emph{International Semantic Web Conference}.\hskip 1em plus 0.5em minus 0.4em\relax Springer, 2022, pp. 39--56.

\bibitem{vashishth2018reside}
S.~Vashishth, R.~Joshi, S.~S. Prayaga, C.~Bhattacharyya, and P.~Talukdar, ``Reside: Improving distantly-supervised neural relation extraction using side information,'' \emph{arXiv preprint arXiv:1812.04361}, 2018.

\bibitem{bastos2021recon}
A.~Bastos, A.~Nadgeri, K.~Singh, I.~O. Mulang, S.~Shekarpour, J.~Hoffart, and M.~Kaul, ``Recon: relation extraction using knowledge graph context in a graph neural network,'' in \emph{Proceedings of the Web Conference 2021}, 2021, pp. 1673--1685.

\bibitem{verlinden2021injecting}
S.~Verlinden, K.~Zaporojets, J.~Deleu, T.~Demeester, and C.~Develder, ``Injecting knowledge base information into end-to-end joint entity and relation extraction and coreference resolution,'' \emph{arXiv preprint arXiv:2107.02286}, 2021.

\bibitem{jain2024revisiting}
M.~Jain, R.~Mutharaju, R.~Kavuluru, and K.~Singh, ``Revisiting document-level relation extraction with context-guided link prediction,'' in \emph{Proceedings of the AAAI Conference on Artificial Intelligence}, vol.~38, no.~16, 2024, pp. 18\,327--18\,335.

\end{thebibliography}


\end{document}